# EyeWorld: A Generative World Model of Ocular State and Dynamics

Ziyu Gao[1], Xinyuan Wu[1], Xiaolan Chen[1], Zhuoran Liu[1], Ruoyu Chen[1], Bowen Liu[1], Bingjie Yan[1], Zhenhan Wang[1], Kai Jin[2,3], Jiancheng Yang[4,5], Yih Chung Tham[6], Mingguang He[1,7,8*], Danli Shi[1,8,*]

[1] School of Optometry, The Hong Kong Polytechnic University, Hong Kong.

[2] Eye Center of Second Affiliated Hospital, School of Medicine, Zhejiang University, Hangzhou, 310009, Zhejiang, China

[3] Zhejiang Provincial Key Laboratory of Ophthalmology, Zhejiang Provincial Clinical Research Center for Eye Diseases, Zhejiang Provincial Engineering Institute On Eye Diseases, Hangzhou, 310009, Zhejiang, China

[4] ELLIS Institute Finland, Espoo 02150, Finland.

[5] Aalto University, Espoo 02150, Finland.

[6] Yong Loo Lin School of Medicine, National University of Singapore, Singapore.

[7] Centre for Eye and Vision Research (CEVR), 17W Hong Kong Science Park, Hong Kong.

[8] Research Centre for SHARP Vision (RCSV), The Hong Kong Polytechnic University, Hong Kong.

* Corresponding authors: danli.shi@polyu.edu.hk, mingguang.he@polyu.edu.hk

**Abstract**

Ophthalmic decision-making depends on subtle lesion-scale cues interpreted across multimodal imaging and over time, yet most medical foundation models remain static and degrade under modality and acquisition shifts. Here we introduce EyeWorld, a generative world model that conceptualizes the eye as a partially observed dynamical system grounded in clinical imaging. EyeWorld learns an observation-stable latent ocular state shared across modalities, unifying fine-grained parsing, structure-preserving cross-modality translation and quality-robust enhancement within a single framework. Longitudinal supervision further enables time-conditioned state transitions, supporting forecasting of clinically meaningful progression while preserving stable anatomy. By moving from static representation learning to explicit dynamical modeling, EyeWorld provides a unified approach to robust multimodal interpretation and prognosis-oriented simulation in medicine.

**Introduction**

Vision impairment affects more than 2.2 billion people worldwide and remains a major cause of disability and reduced quality of life. Most cases arise from common ocular diseases—including cataract, age-related macular degeneration (AMD), glaucoma and diabetic retinopathy (DR)—many of which are preventable or treatable if detected early.[1] Yet early detection and sustained management are unevenly delivered, constrained by fragmented care pathways, limited specialist access and growing clinical demand. As populations age and the prevalence of chronic disease increases, the demand for scalable, reliable ophthalmic assessment continues to grow.[2]

At the same time, ophthalmology has become one of the most imaging-intensive fields in medicine. Routine care integrates colour fundus photography (CFP), optical coherence tomography (OCT) and angiographic imaging, generating detailed structural and vascular information at unprecedented scale. These modalities provide complementary views of the same biological system, capturing both anatomy and pathology across time.[3] However, the volume, heterogeneity and longitudinal complexity of these data increasingly exceed the interpretive capacity of clinicians. The central challenge is no longer data acquisition, but coherent integration.

Artificial intelligence has achieved expert-level performance in well-defined ophthalmic tasks such as disease classification, lesion segmentation and image quality assessment.[4,5] Yet most existing systems are narrow and static. They are trained for a single objective, modality or dataset, and operate in isolation from one another.[6] Medical reasoning is fundamentally different. Clinicians synthesize multimodal evidence, distinguish artefact from pathology, and interpret disease as an evolving biological process rather than a single image to be labelled.[7] A retinal scan is not merely a pattern to classify; it is a partial observation of a dynamic system unfolding over time.

Recent medical foundation models attempt to unify tasks through large-scale representation learning, improving generalization across datasets and objectives.[8-11] However, these approaches remain predominantly representation-centric.[12] They assume that sufficiently powerful encoders can capture clinical intelligence from static observations. In ophthalmology, this assumption reveals important limitations. First, diagnostic meaning often resides in subtle lesion-level structures and their spatial relationships, which may be diluted by coarse objectives. Second, disease progression is inherently dynamical, yet most models treat images as independent samples. Moreover, clinical observations are unstable: identical pathology may appear differently across modalities and devices, while benign variability may mimic progression.[13,14] Effective clinical reasoning therefore requires a latent representation that is invariant to observational shifts but sensitive to biologically meaningful change—and capable of evolving over time.

World models offer a principled alternative. Rather than encoding images solely as features, a world model seeks to learn the latent state of a system and the rules governing its transitions under partial observability.[15] This paradigm has reshaped reinforcement learning and is beginning to influence medical modeling, yet existing medical implementations are typically restricted to single modalities or narrow tasks.[7] A unified, multimodal and longitudinal world model tailored to the complexities of ophthalmic disease has not been established.

Here we introduce EyeWorld, a generative multimodal world model that conceptualizes the eye as a partially observed dynamical system grounded in clinical imaging. EyeWorld learns a shared latent ocular state constrained simultaneously by complementary imaging modalities, allowing fine-grained anatomical delineation, structure-preserving cross-modality translation and acquisition-robust enhancement within a single framework. By incorporating longitudinal supervision, the model also learns time-conditioned state transitions, enabling forward simulation of disease-relevant change across clinically meaningful intervals.

We evaluate EyeWorld across multimodal benchmarks encompassing anatomical parsing, lesion localization, modality translation and image enhancement, and directly test its ability to model temporal dynamics through longitudinal prediction. To probe whether the model captures underlying disease mechanisms rather than superficial appearance patterns, we further perform counterfactual progression synthesis in which pathological factors are selectively perturbed while unrelated anatomy is preserved. These results support a central proposition: that modeling the eye as a shared latent dynamical system that constrained by complementary multimodal observations enables observation-invariant reasoning and forward simulation of disease progression within a unified generative framework. By moving beyond static representation learning toward explicit dynamical modeling, EyeWorld represents a step toward computational systems that reason about disease rather than simply recognize it.

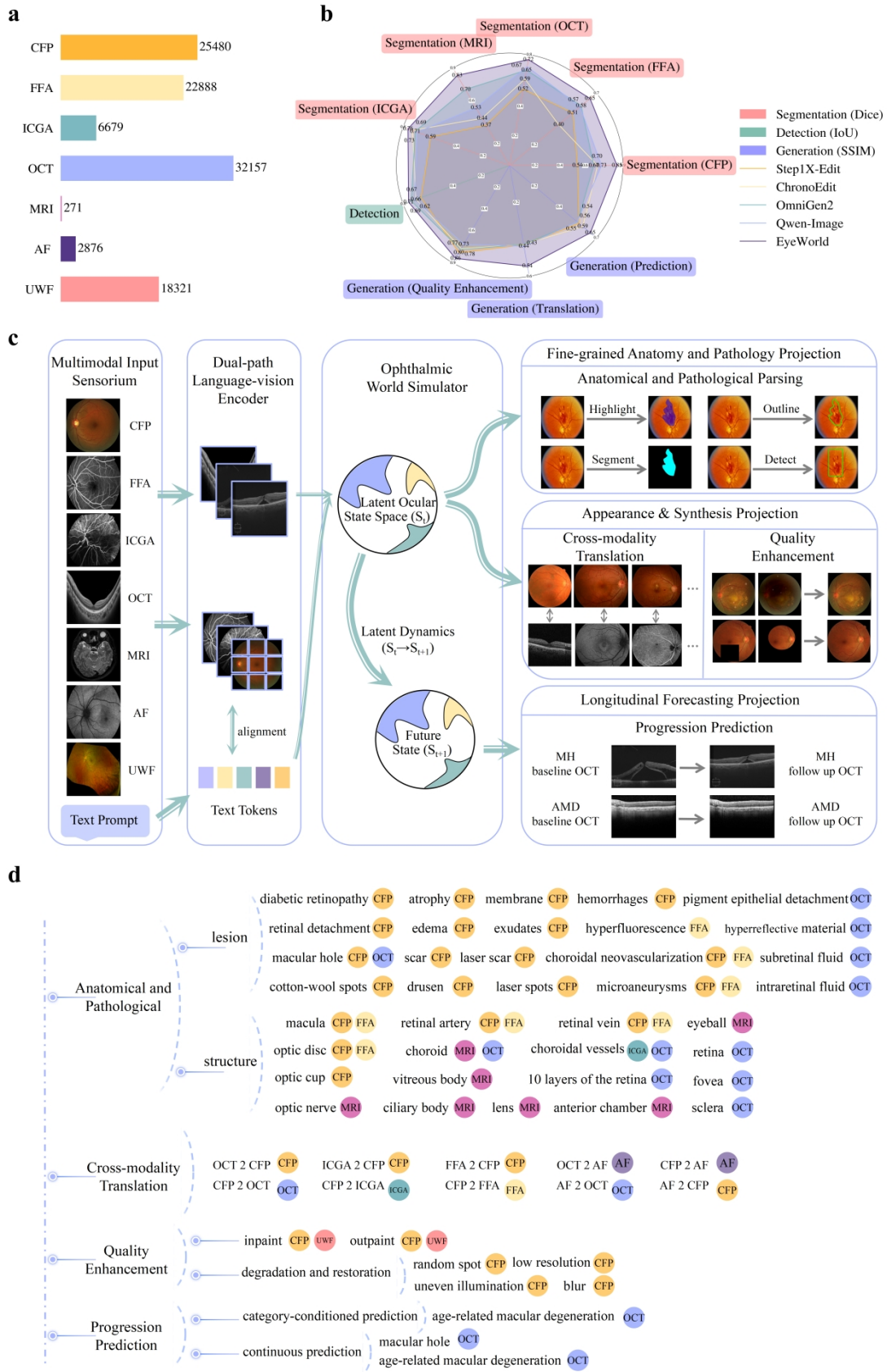

**Fig. 1 | Study overview, data composition and task taxonomy**. **a**, Number of images across modalities in the multimodal ophthalmic dataset, comprising 108,672

images spanning colour fundus photography (CFP), fundus fluorescein angiography (FFA), indocyanine green angiography (ICGA), optical coherence tomography (OCT), magnetic resonance imaging (MRI), autofluorescence (AF) and ultra-widefield (UWF) imaging. **b**, Radar plot comparing EyeWorld with four baseline models across major task families. Segmentation performance is evaluated using Dice similarity coefficient, detection using Intersection over Union (IoU), and generative tasks using structural similarity index (SSIM). Higher values indicate better performance across all metrics. **c**, Schematic overview of EyeWorld. Multimodal images and free-form text prompts are tokenized and fused through a dual-path language-vision encoder into a shared latent embedding space. The Ophthalmic World Simulator operates on this latent state to model diverse clinical objectives. For temporal forecasting, the model learns time-conditioned latent state transitions rather than static pattern mappings, enabling simulation of disease evolution. A unified generative projection head decodes the latent state to support anatomical parsing (segmentation), structure-preserving cross-modality synthesis (translation), and longitudinal forecasting (progression prediction) within a single framework. **d**, Taxonomy of supported tasks. EyeWorld encompasses fine-grained anatomical segmentation, detection of heterogeneous pathological lesions (including macular hole (MH) and age-related macular degeneration (AMD)), and complex generative tasks such as longitudinal disease progression forecasting.

## Results

### Overview of EyeWorld

To develop and evaluate EyeWorld as a generative world model of ocular state and dynamics, we curated a large-scale multimodal ophthalmic dataset comprising CFP, OCT, angiographic modalities, magnetic resonance imaging (MRI), autofluorescence imaging (AF), and ultra-widefield (UWF) imaging. The dataset includes 85,638 image-mask pairs covering 54 anatomical structures and pathological lesion types, 31,179 paired samples for cross-modality translation, and 16,645 longitudinal examinations from pre-operative, post-operative, and follow-up visits (Fig. 1a).

We compared EyeWorld with four representative instruction-guided image generation and editing models: Step1X-Edit[16], ChronoEdit[17], OmniGen2[18], and Qwen-Image[19]. These baselines represent contemporary paradigms for multimodal instruction-following image transformation, including diffusion decoders conditioned on large language model embeddings, video-interpolation-based editing, unified multimodal transformers, and hybrid semantic–variational latent conditioning. EyeWorld is initialized from the OmniGen2 backbone but is restructured into a domain-specific world model that explicitly learns (i) an observation-stable latent ocular state and (ii) time-conditioned state transitions from longitudinal data (Methods).

All baselines were fine-tuned on identical ophthalmic splits and evaluated using standardized prompts and post-processing pipelines. Across major task families, EyeWorld achieved the most consistent performance gains, improving both discriminative objectives (parsing and localization) and generative transformations within a single framework (Fig. 1b), rather than excelling on isolated tasks.

We structured evaluation around three central challenges that constrain current ophthalmic AI systems: (i) preserving fine-grained anatomical and pathological semantics across modalities, (ii) generating structurally faithful images under substantial devices and acquisition variability, and (iii) predicting clinically meaningful longitudinal change. The first two interrogate the stability and fidelity of the learned ocular state; the third evaluates learned ocular dynamics.

**Preserving fine-grained anatomy and pathology across modalities**

We first examined whether EyeWorld preserves clinically critical lesion-level semantics within a shared latent state, rather than capturing only coarse organ-level structure. We constructed a multimodal segmentation test set spanning CFP, FFA, ICGA, OCT and MRI (Fig. 2a). In total, 60 segmentation tasks were evaluated using 12,845 image-mask pairs, covering 29 anatomical structures (including retina, optic disc, fovea, macula, lens and choroidal layers) and 25 lesion types (including haemorrhages, microaneurysms, macular holes and pigment epithelial detachments). All models were prompted to output colour-coded masks, which were converted into binary or multi-class maps via a unified parsing pipeline prior to Dice and mean intersection-over-union (mIoU) computation.

Across all tasks, EyeWorld achieved a mean Dice score of 0.77, outperforming Step1X-Edit (0.52), ChronoEdit (0.61), OmniGen2 (0.64), and Qwen-Image (0.69). Performance gains were consistent across modalities, with mean Dice improvements of 0.19 (CFP), 0.135 (FFA), 0.12 (OCT), 0.32 (MRI) and 0.08 (ICGA) over the baseline average. Stratified analyses confirmed that EyeWorld ranked highest for both anatomical structures and lesions across all modalities (Fig. 2b,c). Notably, improvements were observed across imaging modalities governed by distinct physical principles. On OCT, EyeWorld better delineated thin retinal layers and membranous structures. On angiographic modalities, it more accurately captured vascular topology and leakage regions. On MRI, it produced globally coherent ocular boundaries with higher structural fidelity. This cross-modality consistency suggests that EyeWorld learns a shared latent representation that abstracts anatomical-pathological structure beyond modality-specific appearance cues.

All models exhibited lower performance for lesions than for anatomical structures, reflecting smaller size, irregular geometry and lower contrast. However, EyeWorld substantially reduced this gap. Across modalities, it achieved a mean Dice of 0.84 for anatomical structures and 0.66 for lesions, compared with 0.76 and 0.58, respectively, for the strongest baseline. In CFP specifically, EyeWorld achieved a Dice score of 0.858 for anatomical structures, outperforming OmniGen2 by 8.72% and Step1X-Edit, ChronoEdit, and Qwen-Image by 31.82%, 15.25%, and 11.28%, respectively. For CFP lesion segmentation, EyeWorld reached a Dice of 0.71, exceeding Qwen-Image by 5% and Step1X-Edit, ChronoEdit, and OmniGen2 by 9%, 4%, and 2% (all $P < 0.05$). These gains likely reflect two architectural factors. First, deep language-vision fusion enables lesion-specific prompts (e.g., "segment microaneurysms") to modulate the latent state toward small, clinically meaningful targets. Second, coarse-to-fine training encourages simultaneous preservation of global vascular topology and high-frequency lesion boundaries, which are often smoothed out in general-purpose generators.

To evaluate robustness under structural ambiguity, we constructed segmentation scenarios in which a single prompt corresponded to multiple plausible masks due to overlapping or poorly defined boundaries[20]. Such ambiguity is clinically relevant, for example, in glaucoma assessment, small variations in optic disc and cup delineation can meaningfully alter the cup-to-disc ratio[21]. In this setting, EyeWorld achieved a Dice of 0.865 and mIoU of 0.764, outperforming Step1X-Edit (Dice 0.512), ChronoEdit (0.625), OmniGen2 (0.778) and Qwen-Image (0.695). Beyond higher mean accuracy, EyeWorld exhibited greater boundary stability, indicating that its latent state remains coherent when uncertainty arises from anatomy rather than noise.

To disentangle modality-driven visibility from prompt bias, we evaluated targets visible across multiple modalities using identical modality-agnostic prompts (Fig. 2e). Performance differences aligned with imaging physics rather than textual cues. For macular hole, segmentation accuracy was higher on OCT than on CFP across all models, consistent with OCT's depth-resolved visualization of foveal defects[22]; EyeWorld remained best-performing in both modalities. For optic disc and choroidal vessels, modality-dependent trends also reflected known visibility patterns[23], while EyeWorld consistently ranked highest or near-highest. These findings indicate that the latent state encodes modality-specific visibility grounded in imaging physics rather than overfitting to prompt semantics.

Qualitative comparisons corroborated quantitative results (Fig. 3). Step1X-Edit frequently under-segmented thin or low-contrast structures. ChronoEdit produced over-smoothed or spatially misaligned masks. OmniGen2 captured large regions but fragmented small lesions and introduced irregular contours. Qwen-Image improved semantic alignment but generated scattered false positives for small targets. In contrast, EyeWorld generated contiguous choroidal layers in OCT, precisely localized optic discs in angiography, accurately outlined laser spots in CFP, and produced vascular masks in ICGA that preserved tree-like topology without speckle noise. Across diverse prompts and modalities, EyeWorld adhered to colour conventions and maintained coherent masks across scales.

Together, these findings demonstrate that EyeWorld preserves lesion-scale detail and modality-specific visibility cues within a unified latent ocular state. The same instruction yields modality-consistent yet physics-aware outputs, supporting the conclusion that EyeWorld embeds heterogeneous ophthalmic observations into a stable predictive state space rather than relying on superficial prompt conditioning.

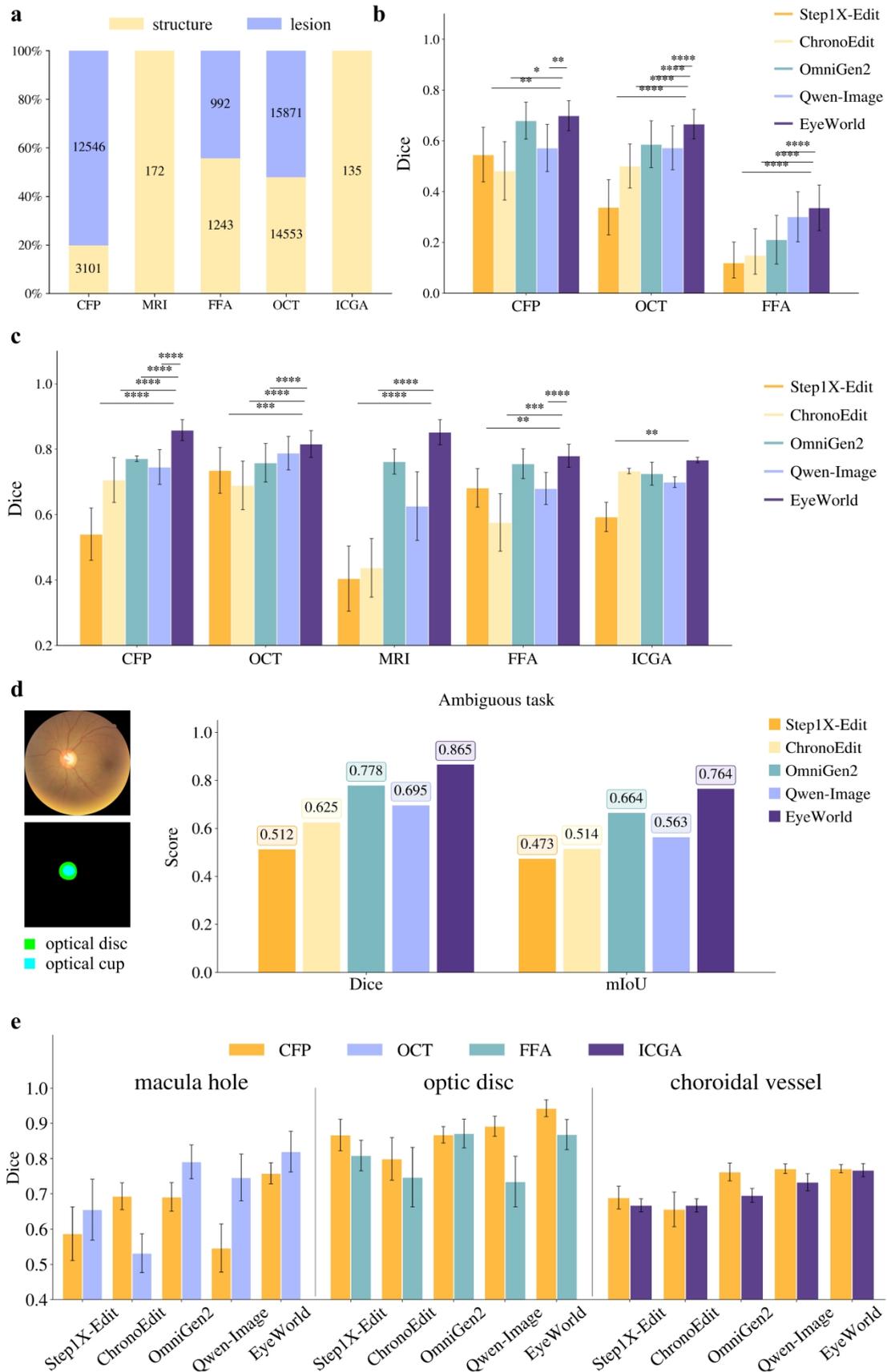

**Fig. 2 | Multimodal anatomical and pathological parsing**. **a**, Distribution of segmentation supervision across imaging modalities, indicating the number of

annotated anatomical structures and pathological lesion categories per modality. **b,c**, Mean Dice scores for lesion (b) and anatomical-structure (c) segmentation across modalities for EyeWorld and baseline models. Error bars denote mean ± s.d. Statistical significance was assessed using two-sided Mann-Whitney U tests (*$P < 0.05$, **$P < 0.01$, ***$P < 0.001$, ****$P < 0.0001$). **d**, Segmentation performance for optic disc and optic cup under structurally ambiguous conditions, evaluating robustness to boundary uncertainty. **e**, Cross-modality segmentation using identical, modality-agnostic prompts. Performance differences reflect modality-dependent visibility for macular hole, optic disc and choroidal vessel targets.

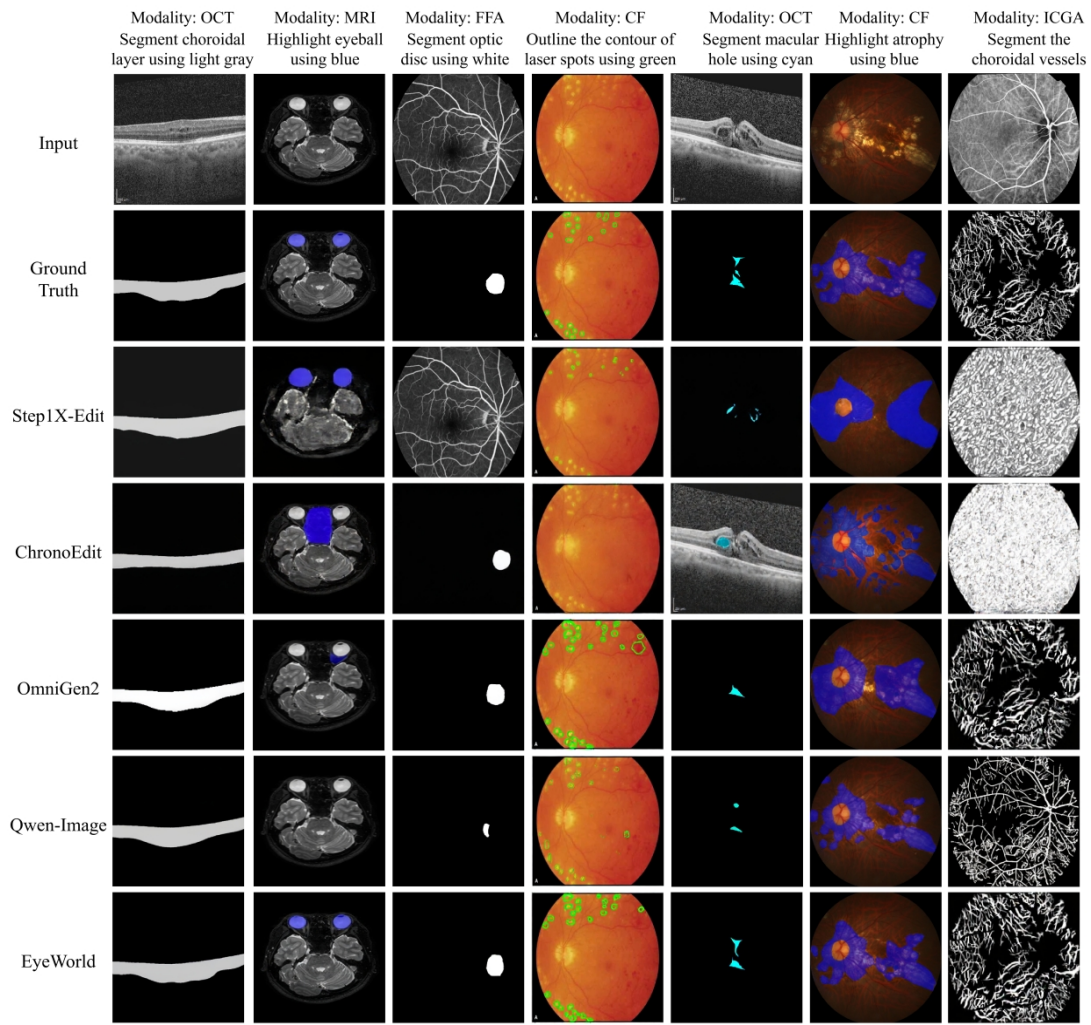

**Fig. 3 | Representative prompt-guided segmentation outputs across modalities.** Illustrative examples of segmentation, highlighting and contour-outlining tasks in optical coherence tomography (OCT), magnetic resonance imaging (MRI), fundus fluorescein angiography (FFA), colour fundus photography (CFP) and indocyanine green angiography (ICGA). Each column corresponds to a text prompt specifying the target structure or lesion and its required colour convention. Rows show the input image, ground-truth mask, and outputs generated by Step1X-Edit, ChronoEdit, OmniGen2, Qwen-Image and EyeWorld.

**Robust anatomy and pathology localisation**

We next assessed whether the shared latent ocular state supports object localisation when outputs are bounding boxes rather than pixel-level masks, as required for rapid screening and clinical triage[24]. The detection benchmark comprised 14 tasks across structures and lesions, summarized by mean Intersection over Union (IoU) distributions (Fig. 4a). EyeWorld achieved the highest number of tasks with mean IoU > 0.7 (four tasks), compared with none for Step1X-Edit, one for ChronoEdit, three for OmniGen2 and one for Qwen-Image. Concurrently, EyeWorld reduced the number of tasks with mean IoU < 0.5 to six, whereas baseline models placed between seven and ten tasks in this lower-accuracy regime.

Qualitative comparisons revealed distinct and complementary failure modes among baselines (Fig. 4b). Step1X-Edit occasionally failed to generate any bounding box despite correct prompts, indicating brittle adherence to structured output constraints. ChronoEdit frequently produced a single conservative box capturing only the most salient portion of diffuse pathology, such as geographic atrophy, thereby underestimating spatial extent. OmniGen2 generated spatially plausible boxes but sometimes drifted toward high-contrast background regions, suggesting reliance on appearance-driven cues. Qwen-Image often output multiple fragmented boxes, increasing recall at the expense of precision through false-positive detections. In contrast, EyeWorld more accurately captured the full spatial extent of diffuse atrophy and more reliably isolated sparse haemorrhages, reflecting stronger spatial grounding in the underlying anatomy.

Aggregated across tasks, EyeWorld achieved the highest mean IoU and precision for both lesion and structural detection (Fig. 4c; shaded areas indicate mean ± s.d.). These findings demonstrate that the learned latent representation is not tied to a specific output format. A single world-model state supports both fine-grained delineation and coarse localisation without task-specific architectural branches. This flexibility is consistent with the objective of mechanistic abstraction: spatial structure encoded in the latent state can be decoded into different clinically relevant outputs while preserving anatomical and pathological coherence.

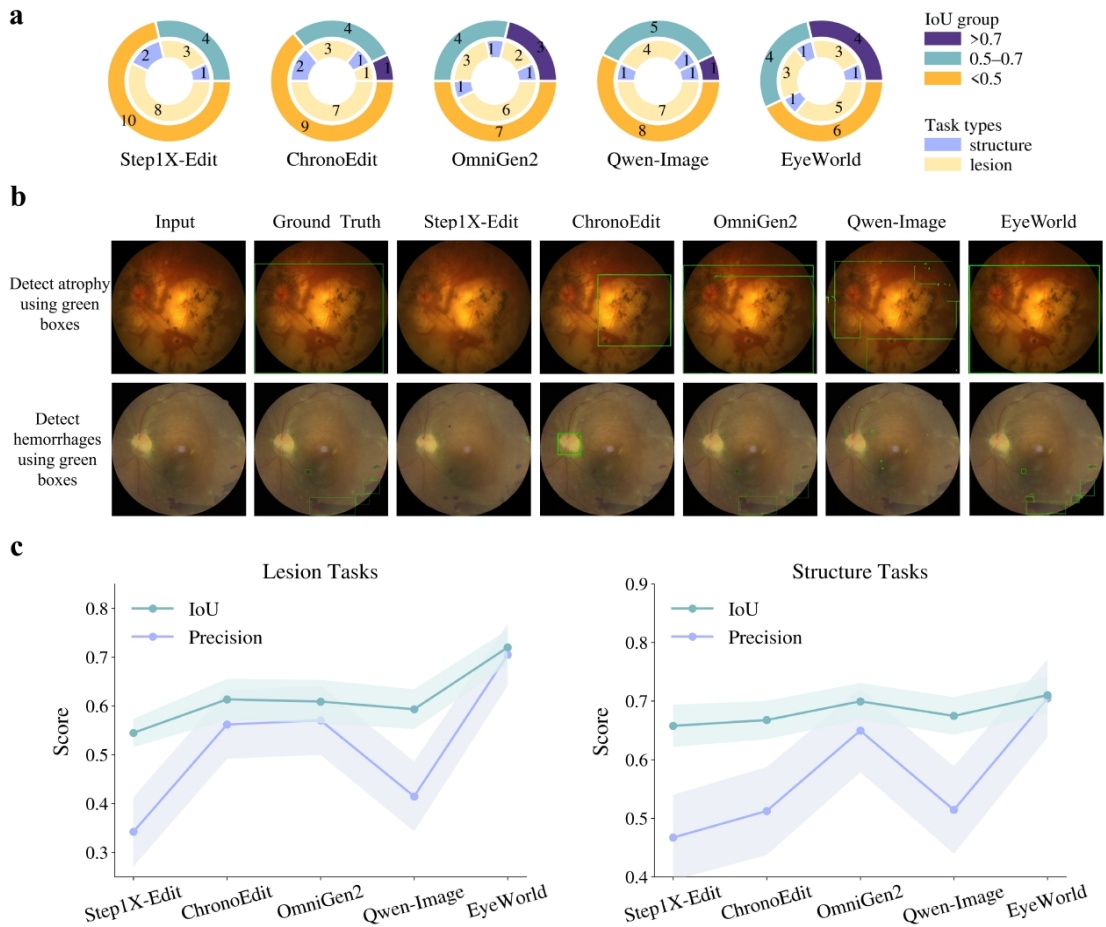

**Fig. 4 | Lesion and structure localisation performance. a**, Distribution of 14 detection tasks stratified by mean Intersection over Union (IoU) for each model, grouped into IoU > 0.7, 0.5–0.7 and < 0.5. **b**, Representative detection examples in colour fundus photography (CFP), showing predicted bounding boxes alongside ground-truth annotations. **c**, Aggregate IoU and precision for lesion and anatomical structure detection across all tasks. Shaded areas indicate mean ± s.d.

**Structure-preserving cross-modality translation**

Comprehensive ophthalmic assessment often requires multiple imaging modalities, yet multimodal acquisition increases patient burden and may be impractical in resource-limited settings[10,25]. Cross-modality translation therefore serves both as a clinically relevant modality-completion task and as a stringent test of domain heterogeneity: the model must preserve retinal topology while transforming modality-specific imaging physics, contrast mechanisms and texture statistics.

We constructed paired translation benchmarks across five ophthalmic modalities using spatially aligned images from the same eye, with prompts explicitly specifying the target modality (for example, "translate CFP to late-phase FFA") (Fig. 5a). For each task, paired images provided pixel-aligned supervision. Image-level fidelity was evaluated using PSNR, SSIM and LPIPS, and distribution-level realism and diversity were assessed using Inception Score (IS) and Perceptual Image Distance (PID)[26,27].

Baseline models exhibited characteristic inductive biases. Step1X-Edit preserved coarse fundus layout but attenuated fine vessels and optic disc detail when translating AF to CFP, suggesting limited joint modelling of modality-specific texture and structure. ChronoEdit frequently produced overly smooth outputs with reduced contrast, particularly detrimental in angiographic targets where capillary-level detail is diagnostically critical. OmniGen2 occasionally introduced fluorescence-like artefacts unrelated to vasculature, consistent with hallucinations under domain shift from natural-image priors. Qwen-Image generated sharper images but often over-amplified bright lesions or background granularity, leading to unrealistic speckle patterns in late-phase FFA or ICGA. In contrast, EyeWorld consistently preserved vessel calibre, optic disc geometry and global illumination while accurately reproducing modality-specific signatures, including fluorescence distribution in angiography and retinal layer stratification in OCT (Fig. 5a).

Quantitatively, EyeWorld achieved the most favourable distributions across LPIPS, PSNR and SSIM (Fig. 5b), indicating lower perceptual distance and higher structural fidelity relative to all baselines. Across 18 translation directions, EyeWorld consistently produced higher IS and lower PID (Fig. 5c). For example, in CFP→FFA translation, EyeWorld improved IS by 4.05% over the strongest baseline; in ICGA→CFP translation, the improvement reached 8.12%. Lower PID values indicate closer alignment with target-domain distributions in deep feature space.

Together, these results indicate that EyeWorld learns a stable shared generative space in which heterogeneous ophthalmic modalities can be projected without disrupting retinal topology. Clinically, such structure-preserving translation could support modality imputation, reduce reliance on invasive imaging, and extend multimodal assessment to settings where specific modalities are unavailable.

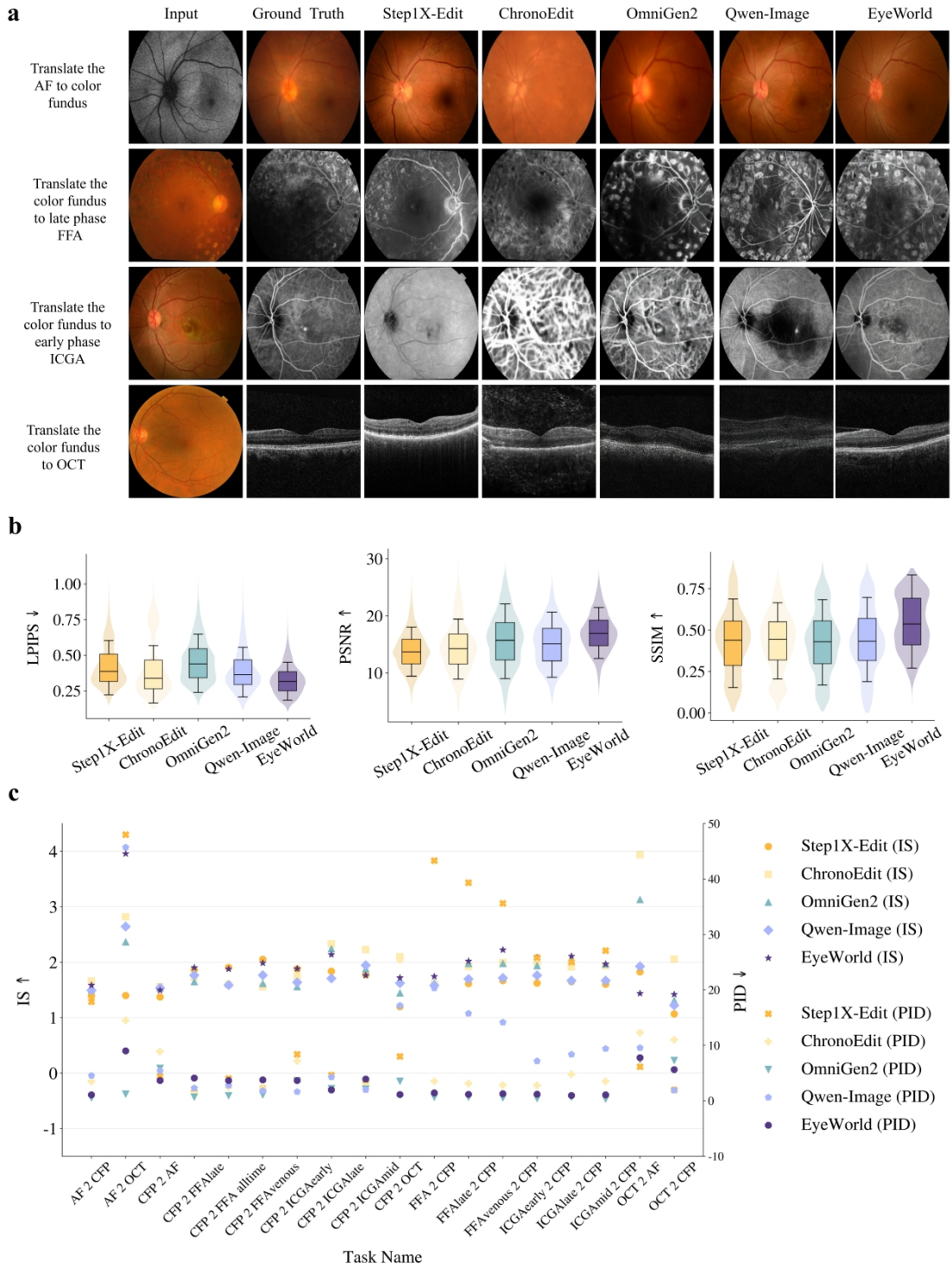

**Fig. 5 | Cross-modality translation performance**. **a**, Representative translations across modality pairs, showing source image, paired reference and model outputs. **b**, Distributions of LPIPS, PSNR and SSIM across translation tasks; lower LPIPS and higher PSNR/SSIM indicate better image-level fidelity. **c**, Direction-wise comparison of distribution-level realism and diversity using Inception Score (IS) and Perceptual Image Distance (PID); higher IS and lower PID indicate closer alignment with the target-domain distribution.

**Image quality enhancement under acquisition heterogeneity**

To evaluate robustness under real-world acquisition variability, we framed image quality enhancement as observation recovery: the model should correct device-dependent degradations while preserving the underlying ocular state, including lesion geometry and retinal topology. We evaluated four challenging scenarios: combined degradation restoration[28], super-resolution[29,30], inpainting and outpainting[31,32] (Fig. 6a). These scenarios emulate common clinical artefacts, including blur, reduced resolution, uneven illumination, local corruption and truncated fields of view, all of which can obscure lesions and compromise interpretation.

Baseline models showed distinct failure modes (Fig. 6a). Under combined degradation, Step1X-Edit often left residual bright circular artefacts, indicating limited modelling of artefact formation. ChronoEdit removed artefacts but introduced contrast drift and colour bias. OmniGen2 reduced noise yet produced a persistent haze that softened vessel boundaries. Qwen-Image sharpened structures but over-smoothed subtle textures, diminishing small-vessel visibility. In inpainting, ChronoEdit and OmniGen2 blurred across masked regions, disrupting layer continuity, whereas Qwen-Image introduced intensity mismatches at mask seams. In outpainting, OmniGen2 frequently generated peripheral streaks or repetitive patterns, reflecting generic diffusion priors under large extrapolation. EyeWorld more consistently restored illumination and vessel trajectories under degradation, reconstructed high-frequency detail in super-resolution without ringing artefacts, and completed missing regions with anatomically coherent continuity in both inpainting and outpainting.

Quantitatively, EyeWorld achieved the lowest LPIPS across all four task categories, with statistically significant pairwise improvements (Fig. 6b). LPIPS values were 0.142 (combined degradation), 0.170 (super-resolution), 0.215 (inpainting) and 0.301 (outpainting). In combined degradation, EyeWorld outperformed Step1X-Edit, ChronoEdit, OmniGen2 and Qwen-Image by margins of 0.208, 0.213, 0.246 and 0.265, respectively.

These results demonstrate that EyeWorld's latent state supports structure-preserving generation under heterogeneous acquisition noise, enabling enhancement and completion without task-specific specialization.

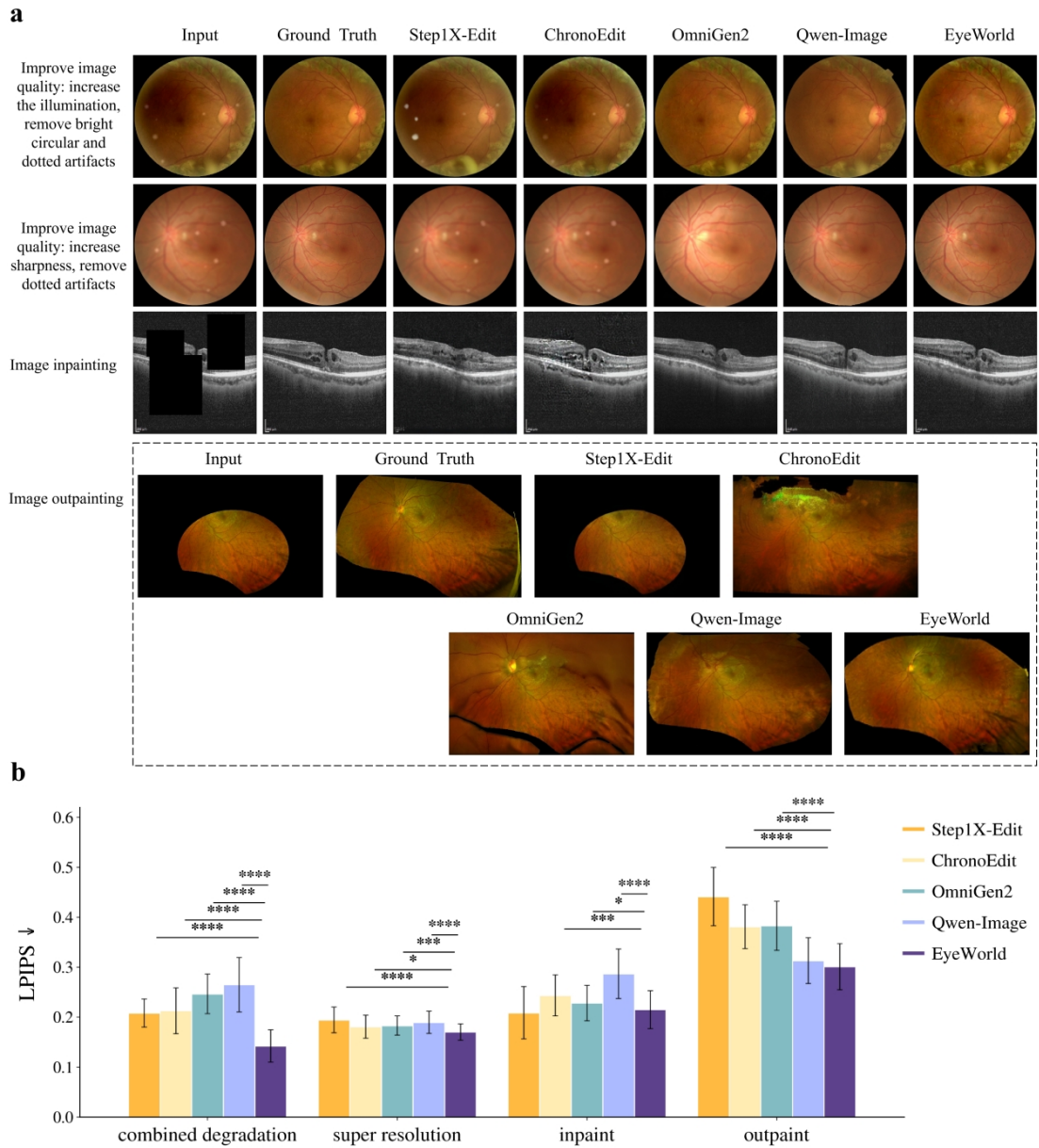

**Fig. 6 | Image enhancement and completion under simulated degradations**. **a**, Representative examples for combined degradation restoration, super-resolution, inpainting and outpainting, comparing model outputs with paired reference images. **b**, Mean LPIPS across task categories; lower values indicate closer perceptual similarity to ground truth.

## Predicting clinically meaningful disease progression

We next evaluated whether EyeWorld captures temporal dynamics through longitudinal OCT prediction in cohorts with age-related macular degeneration (AMD) and macular hole, spanning both gradual degenerative change and pronounced structural transitions. Predicting future retinal appearance is central to longitudinal management, as follow-up schedules and treatments are guided by anticipated trajectories rather than single-visit snapshots[33,34]. Progression forecasting therefore directly tests whether the latent ocular state can be advanced in time to simulate disease evolution.

We focused on OCT datasets with wide follow-up intervals, ranging from weeks to several years (Fig. 7a,d). In AMD, two complementary settings were evaluated (Fig. 7a). First, continuous forecasting conditioned on baseline OCT, patient metadata (age and sex) and follow-up interval (Fig. 7b). Second, category-conditioned prediction reflecting clinically meaningful outcomes—stable (71.3%), recovery (16.3%) and progression (12.3%) (Fig. 7c). For macular hole postoperative recovery, an RPE layer mask was used to enforce spatial alignment between baseline and follow-up scans (Fig. 7g), ensuring evaluation emphasized anatomical change rather than scan positioning variability.

Across all prediction metrics, EyeWorld achieved the lowest LPIPS and highest PSNR and SSIM, with statistically significant improvements (Fig. 7e). Aggregated across tasks, EyeWorld attained a mean LPIPS of 0.371, SSIM of 0.651 and PSNR of 21.462, exceeding the strongest baseline by 2.3%, 6.1% and 2.278, respectively.

Qualitatively, EyeWorld captured both subtle and pronounced morphological changes while preserving global retinal geometry (Fig. 7f,g). In AMD forecasting, Step1X-Edit produced overly smooth follow-ups that under-represented drusen-associated deformation. ChronoEdit introduced shape changes inconsistent with baseline anatomy. OmniGen2 and Qwen-Image matched global contrast yet failed to localize outer-retinal changes. EyeWorld more faithfully reproduced drusen morphology, RPE elevation, outer-retinal remodelling and postoperative restoration of the foveal contour.

By integrating baseline anatomy, patient metadata and follow-up interval, EyeWorld generates temporally coherent and clinically plausible future states. Combined with results on parsing, localisation, translation and enhancement, these findings demonstrate that EyeWorld abstracts multimodal ophthalmic observations into a unified latent state that preserves fine-grained structure while remaining temporally predictive over extended follow-up intervals.

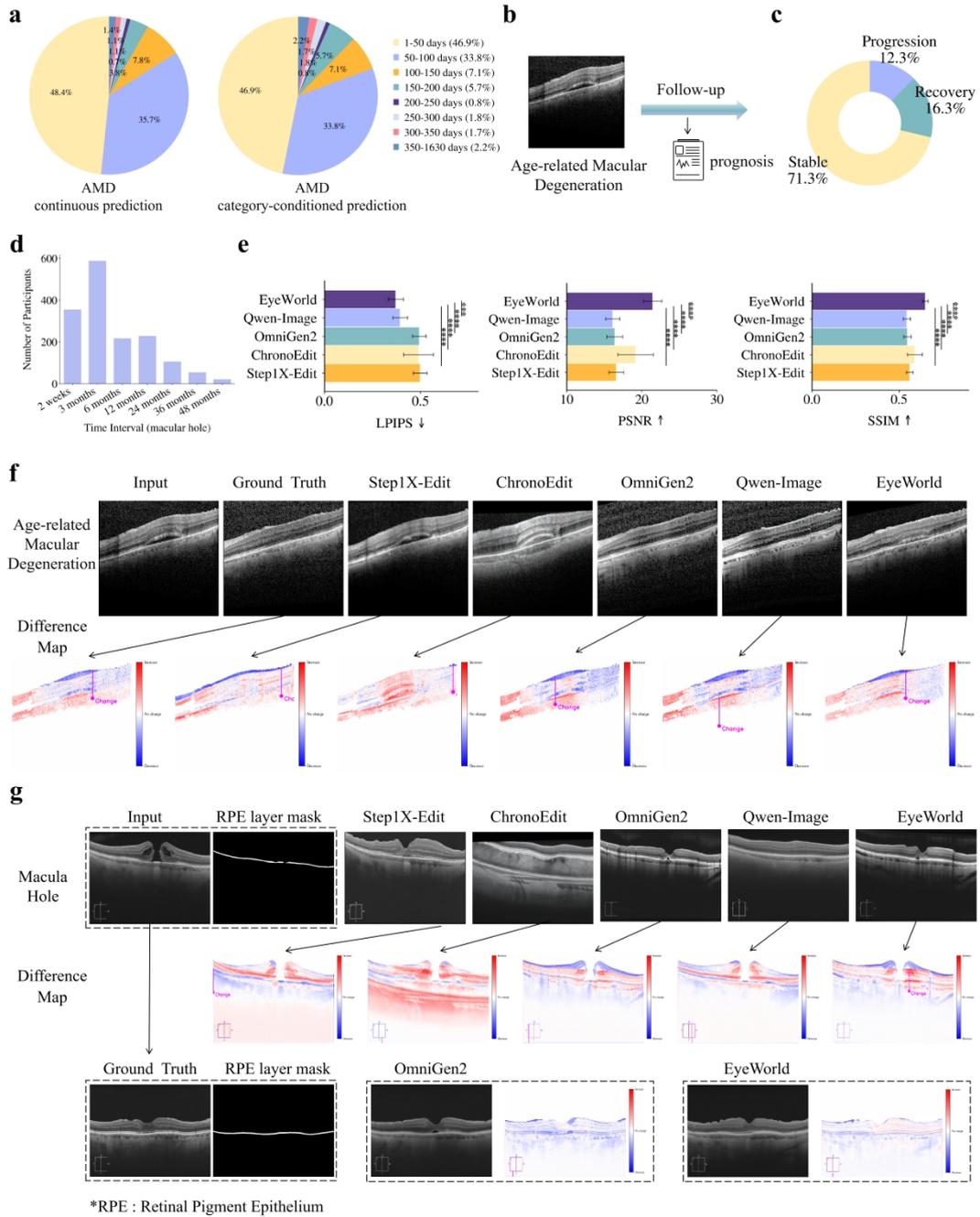

Fig. 7 | **Retinal disease progression prediction performance. a**. Follow-up interval distributions for AMD continuous and category-conditioned prediction settings. **b**, Schematic of progression forecasting conditioned on baseline OCT and prompt-specified follow-up information. **c**, Outcome category distribution for AMD prediction (stable, recovery, progression). **d**, Distribution of macular hole follow-up intervals. **e**, Quantitative comparison across prediction tasks using LPIPS, PSNR and SSIM; lower LPIPS and higher PSNR/SSIM indicate better fidelity. **f**, Representative continuous AMD prediction examples with baseline-to-predicted difference maps; arrow indicates maximal tissue-masked absolute intensity change. **g**, Representative macular hole postoperative recovery prediction using RPE mask alignment, with difference maps highlighting structural changes. **Abbreviations:** AMD, age-related

macular degeneration; RPE, retinal pigment epithelium

**Extended evaluations: robustness and exemplar-conditioned control**

Beyond the core benchmarks, we performed three extended evaluations to test whether EyeWorld's latent ocular state remains reliable under clinically realistic perturbations, supports controllable state evolution, and provides a unified task interface (Fig. 8).

We first evaluated segmentation under a controlled quality shift with three input conditions: degraded quality, original quality, and an "enhance-then-analyse" setting where degraded inputs were first restored and then segmented. Across all conditions, EyeWorld remained the best-performing model (Fig. 8a). Under degraded quality, EyeWorld achieved a Dice score of 0.722, compared with 0.372 (Step1X-Edit), 0.341 (ChronoEdit), 0.623 (OmniGen2) and 0.459 (Qwen-Image). Under original quality, EyeWorld reached 0.780 versus 0.389, 0.363, 0.747 and 0.606, respectively. Notably, in the enhance-then-analyse setting, segmentation performance decreased for all baselines relative to directly analysing degraded inputs (for example, Step1X-Edit: 0.372→0.183; ChronoEdit: 0.341→0.166; OmniGen2: 0.623→0.502; Qwen-Image: 0.459→0.357), whereas EyeWorld slightly improved (0.722→0.731), indicating that its restoration preserves the structural cues required for downstream delineation rather than introducing distribution shifts that weaken lesion evidence.

We next assessed whether EyeWorld can enact targeted progression change without globally altering anatomy by prompting counterfactual follow-ups on AMD OCT. Specifically, given a baseline scan and a stable reference follow-up, we prompted the model to generate a worsened follow-up over the same horizon. As illustrated in Fig. 8b, EyeWorld produces a counterfactual output whose difference map concentrates changes within pathology-associated regions, rather than exhibiting broad tone or contrast drift. This behaviour suggests that the learned representation can selectively modify disease-relevant factors while keeping unrelated structure stable.

Finally, we evaluated a demonstration-conditioned inference protocol (EyeWorld_exemplar) where the model infers the intended operation from a reference input-output pair and applies it to a new query[35]. Fig. 8c summarises performance under this single interface across modalities and task families, including segmentation (Dice), detection (IoU) and generative tasks (SSIM) spanning translation, quality enhancement and longitudinal prediction. EyeWorld achieves strong segmentation performance across modalities, with Dice scores of 0.79 on CFP, 0.60 on OCT, 0.56 on FFA, 0.53 on MRI and 0.76 on ICGA, indicating that the inferred operation remains stable despite systematic shifts in contrast, resolution and field of view. The same exemplar interface extends to detection, reaching an IoU of 0.73, and to generative tasks evaluated by SSIM, including translation (0.60), quality enhancement (0.77) and longitudinal prediction (0.55). Representative examples are shown in Fig. 8d. When asked to "apply the same process" as the reference pair, EyeWorld correctly transfers a lesion highlighting operation to a new case, localising atrophy while preserving background retinal appearance. For macular hole, it generates a compact, spatially plausible segmentation consistent with the ground truth target size and location. For artery and vein segmentation, the output maintains global vascular topology and preserves fine calibre branches while keeping the artery-vein separation consistent across the field, reflecting structure constrained inference rather than texture matching. Finally, in AF to colour translation, EyeWorld preserves optic disc location and major vessel trajectories while synthesising modality appropriate colour and illumination, demonstrating that exemplar conditioning can control both structured outputs and appearance transformation within one interface.

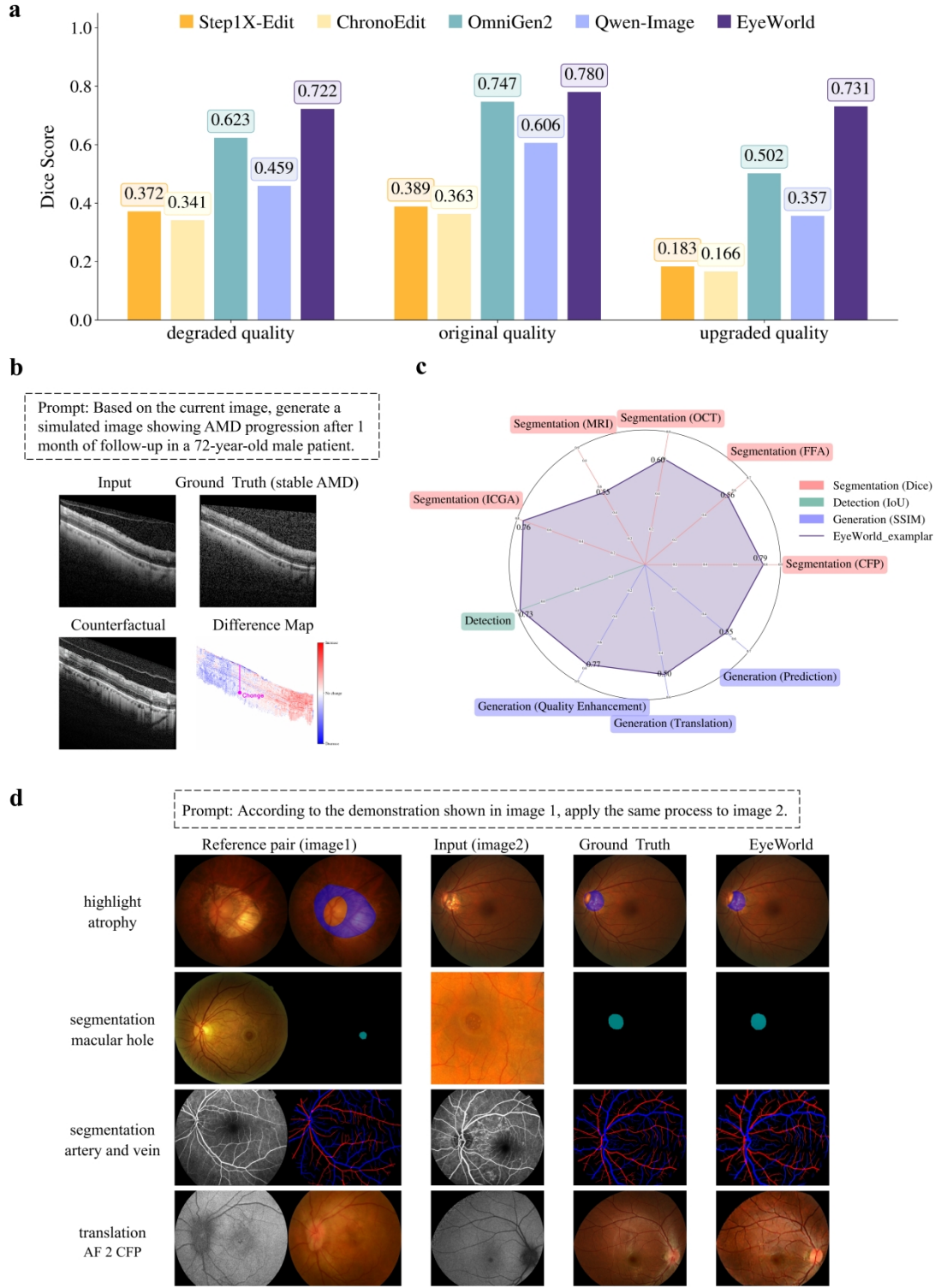

**Fig. 8 | Quality robustness, temporal counterfactual simulation and exemplar conditioned task control of EyeWorld. a**, Segmentation Dice scores under degraded quality, original quality and upgraded quality settings, comparing EyeWorld with Step1X-Edit, ChronoEdit, OmniGen2 and Qwen-Image. **b**, Counterfactual generation for AMD follow up on OCT, showing the input scan, a stable reference, the counterfactual output and the corresponding difference map. **c**, Summary profile of EyeWorld under exemplar based conditioning (EyeWorld_exemplar), reporting

segmentation (Dice) across modalities, detection (IoU), and generative tasks (SSIM) including translation, quality enhancement and longitudinal prediction. **d**, Exemplar conditioned task inference, where EyeWorld learns the intended operation from a reference input output pair and applies it to a new query image, demonstrated for lesion highlighting, segmentation and cross modality translation.

**Discussion**

Ophthalmology provides a stringent testbed for universal medical models. Clinically actionable signals are often subtle and lesion-scale, evolve over time, and must be interpreted across substantial heterogeneity in modality, device and image quality. Guided by these constraints, we developed EyeWorld as a generative world-model–style simulator trained on a curated multimodal dataset that integrates dense structural and lesion-level supervision, paired cross-modality samples and longitudinal follow-up examinations. Across evaluations, EyeWorld exhibits three defining properties consistent with a world-model objective: preservation of lesion-level semantics and boundary-sensitive structure under realistic perturbations; maintenance of observation-consistent anatomy–pathology relationships across modality and acquisition shifts; and time-conditioned forecasting that expresses spatially localized progression while preserving unrelated anatomy. Together, these results suggest that a unified latent ocular state, constrained by complementary clinical readouts and longitudinal supervision, can bridge fine-grained parsing, robust multimodal generation and prognosis-oriented simulation within one framework.

**Observation-stable ocular state under modality and acquisition shifts.** A practical ophthalmic foundation model must treat different acquisitions as partial observations of the same underlying anatomy-pathology configuration, instead of fragmenting its representation by modality, device or quality. EyeWorld advances this objective by coupling structured parsing objectives with generative supervision in a shared latent space. Empirically, this coupling manifests as consistent gains in lesion and structure delineation and localisation across modalities, alongside improved structure-preserving cross-modality translation and restoration. Importantly, the extended robustness analyses further suggest that EyeWorld's gains are not limited to clean in-distribution inputs: when local evidence is attenuated by realistic degradations, the model remains comparatively stable, and enhancement does not undermine downstream delineation. This behaviour is clinically consequential because ophthalmic decision-making often hinges on marginal contour differences and small lesions, where visually plausible outputs that distort lesion geometry can be more harmful than conservative predictions.[36] Conceptually, the results are consistent with treating retinal anatomy as a constrained system: by learning representations that must simultaneously support pixel-level structure, object-level spatial grounding and appearance transformation, the latent state becomes less sensitive to superficial texture statistics that vary across acquisitions.

Heterogeneity also arises from the multi-modality nature of ophthalmic practice,

where complementary modalities are used to interrogate microstructure, vasculature and metabolic signals. However, acquiring all modalities for every patient increases burden, prolongs workflows and may be infeasible in resource-limited settings.[37] Cross-modality translation is therefore clinically meaningful only if it preserves topology and lesion semantics while synthesising modality-appropriate appearance, rather than producing visually plausible but diagnostically misleading outputs. EyeWorld is designed to meet this requirement by learning a shared predictive space that supports both structured parsing and multimodal generation, so that modality transfer is constrained by the same anatomy-aware representation used for analysis (Fig. 5). The consistent gains of EyeWorld across translation and related generative tasks, together with its robustness under quality perturbations, support the view that its outputs remain anchored to retinal geometry rather than drifting with modality- or device-specific superficial cues. In contrast, approaches optimized primarily for generic editing or broad visual realism can be more sensitive to domain shift, for example by smoothing fine lesion boundaries or introducing texture changes that are perceptually plausible but not anatomically faithful.

These properties translate into two deployment-relevant implications. First, reliable modality translation could reduce imaging burden in follow-up workflows by approximating a missing modality from an available acquisition, potentially shortening clinic visits and limiting exposure to invasive procedures when direct acquisition is not feasible. Second, when acquisition quality is imperfect, enhancement that preserves lesion geometry can expand the usable pool of scans for downstream analysis, which is particularly valuable for screening programmes with variable imaging conditions.

**Temporally predictive state evolution for progression modelling.** Ophthalmic care is inherently temporal: clinicians interpret disease by comparing trajectories across follow-up visits rather than judging isolated images. EyeWorld addresses this gap by embedding temporal supervision into its latent space, so that a single latent state jointly encodes the current anatomical-pathological configuration and can be rolled forward to generate a plausible future state over clinically relevant interval. In OCT progression prediction, this formulation enables EyeWorld to preserve global retinal geometry while expressing local morphological evolution, consistent with the world-model objective of learning predictive state variables rather than fitting a static input-output mapping (Fig. 7).

Moreover, counterfactual progression synthesis provides a functional probe of whether the learned representation captures disease-relevant degrees: if worsening can be induced while unrelated anatomy remains stable and changes concentrate in pathology-associated regions, the model is less likely to be performing a generic "style transfer" that alters appearance without respecting biological structure.[38-40] Counterfactual synthesis offers a practical audit of what morphological adjustments the model associates with a change in progression state, especially in early disease where subtle local changes can trigger different monitoring or treatment decisions. More broadly, Fig. 8b supports the second challenge: moving beyond retrospective description toward prognosis-oriented simulation. Instead of only answering "does the current image look diseased?", the model can approximate "if the current trend continues, what structural changes are likely at the next visit?". Clinically, patients predicted to worsen can be prioritized for shorter follow-up intervals or earlier intervention, whereas those predicted to remain stable might safely extend follow-up, improving resource allocation.

**Unified task specification through exemplar-based conditioning.** Beyond improving performance within individual task benchmarks, EyeWorld also addresses the single-task fragmentation by supporting clinically aligned task control through exemplar-based conditioning. In this setting, the prompt provides a reference input-output pair together with a query image, and the model infers the transformation rule from the example and applies it to the query without any task-specific reconfiguration.[35] Practically, EyeWorld can be driven by a minimal instruction that points to the demonstration, for example, "According to the demonstration shown in image 1, apply the same process to image 2." By grounding the operation in an explicit visual before–after example, exemplar conditioning reduces sensitivity to prompt phrasing, resolves ambiguity in task specification, and preserves anatomical constraints by executing the transformation through the same unified latent state that supports structured parsing and generation. Fig. 8c-d show that EyeWorld's controllability is not an auxiliary feature but an operational consequence of learning a shared latent representation that supports task inference, cross modality generalisation and anatomically grounded outputs.

From a deployment perspective, exemplar conditioning provides a scalable route to handle infrequent targets and evolving clinical requests without retraining a dedicated head or redesigning prompts. It also mirrors how clinicians reason in practice, where current examinations are interpreted by comparison to prior cases.[41] By enabling "reference-guided" operations within a single interface, EyeWorld moves closer to a usable clinical assistant that can flexibly perform analysis and synthesis while remaining anchored to anatomically faithful structure. More broadly, exemplar conditioning reinforces the world-model view: if the latent state is truly shared and predictive, then task specification can be treated as conditioning over the same state space, rather than as isolated per-task modules.

Despite the advances presented here, several limitations warrant further investigation. First, class imbalance, particularly when targets differ markedly in size, shape and prevalence, remains challenging despite targeted augmentation. Second, progression prediction currently focuses on OCT cohorts and specific diseases, and broader validation across devices, institutions and additional pathologies is required to establish generalization under real-world shifts. Third, progression prediction and counterfactual evaluation are currently based on imaging endpoints and expert labels. Prospective longitudinal studies linking EyeWorld's predictions to real-world outcomes such as visual acuity changes and treatment response are needed to establish clinical utility. Fourth, like other generative models, EyeWorld may synthesize changes that are not fully supported by the available evidence in the input, particularly for ambiguous cases, rare conditions or underspecified counterfactual prompts. This uncertainty motivates careful use and explicit safeguards when interpreting simulated progression.

In conclusion, EyeWorld is a multimodal ophthalmic world model trained on a wide spectrum of ophthalmic images and validated across multiple clinical scenarios, and it maintains robust performance across diverse tasks and imaging conditions. By jointly enabling anatomically grounded segmentation, lesion-focused detection and temporally informed prognostic forecasting within a single framework, EyeWorld begins to shift the paradigm from simple generation to simulation, from isolated analysis to holistic understanding, and from a generic foundation model toward a world model in ophthalmic AI. Its development strategy, which combines systematic data curation, curriculum learning, in-context learning and rigorous clinical validation, offers a practical template for building medical AI systems that can adapt to varying levels of clinical expertise and heterogeneous healthcare settings. Taken together, these results point to promising directions for extending similar world model frameworks to other medical specialties, with the important caveat that EyeWorld is non-diagnostic and designed to operate under human oversight.

## Methods

### Ethics statement

This study was conducted in accordance with the Declaration of Helsinki and received approval from the Hong Kong Polytechnic University's institutional review board (HSEARS20240202004). The Institutional Review Board waived informed consent due to the retrospective analysis of anonymized ophthalmic images and public datasets.

### Data sources

The EyeWorld corpus was constructed by integrating large-scale, de-identified clinical data from tertiary eye centers with diverse public datasets. Here we describe the data sources and curation procedures. A detailed summary of data source, modalities, annotation targets and dataset roles is provided in Supplementary Table S1.

For anatomical and pathological parsing, we integrated public retinal datasets with pixel-level lesion such as PALM[42], IDRiD[43], MESSIDOR[44], MMAC[45], OIMHS[46]. Additional samples from ODIR-5k[47], OphthalWeChat[48] and private datasets[49,50] spanning CFP, OCT, FFA, ICGA and MRI were manually annotated by clinicians to harmonize lesion definitions and expand disease coverage.

Paired cross-modality datasets were obtained from Thailand (APTOS-2024[51], paired CFP-OCT) and multiple centers across China, including CFP paired with autofluorescence (AF)[52], FFA[53] and ICGA[54]. The translation corpus comprises 2,876 CFP-AF pairs, 1,106 CFP-OCT pairs and 6,544 CFP-ICGA pairs spanning early, mid and late post-injection phases. For CFP-FFA translation, we included 4,927 CFP-FFA pairs (venous/late phases), 4,526 FFA-CFP reverse pairs and 11,200 timestamped CFP-FFA frame pairs (for example, 00:12.50) to enable phase-aware modelling.

Progression supervision was derived from longitudinal OCT cohorts linking baseline to follow-up examinations after intervention or repeated visits. These include a post-vitrectomy macular hole cohort[55] and the MARIO[56] dataset for AMD, which provides categorical progression labels (reduced, stable or worsened). For quality enhancement experiments, CFP and OCT images were sampled from the parsing pool, and ultra-widefield (UWF) images[57] were incorporated to expose the model to variation in device characteristics and field of view while preserving clinically relevant anatomy and pathology.

**Data preprocessing and splitting**

In clinical practice, medical imaging protocols vary among different modalities, thereby leading to the highly variable intensities of medical images. There exists heterogeneity among different datasets, such as inconsistent labelling standards and varying data resolutions, with a large effect on the development of AI-based models. To ensure uniformity and compatibility of collected medical data, we preprocessed all image pairs with a unified preprocessing protocol, which consists of four major steps, including label standardization, unqualified data exclusion and data resizing. First, we employed a standardized labelling protocol to enhance annotation uniformity: (1) multilabel annotations were converted into binary-label annotations by separating foreground targets, ensuring that each mask represents only one target class; and (2) each target was assigned a unified medical terminology to ensure consistent class definitions for identical objects. Furthermore, rigorous inclusion criteria were implemented to maintain data quality. Specifically, we excluded data samples with incomplete images, missing labels or tiny annotated targets (where the sum of target pixel values is less than 50). Finally, for 3D imaging modalities such as OCT and MRI, volumetric scans were decomposed into in-plane 2D slices to enable uniform processing across all modalities.

Each dataset was partitioned into 70% for training, 15% for validation, and 15% for testing. To achieve a mutually exclusive split on a per-patient basis, we assigned a unique identifier to each patient and grouped their images separately. All slices from the same 3D volume and all images from the same patient were assigned to the same subset to prevent information leakage and maintain strict subject-level independence.

**Task formulations and training data construction**

**Anatomical and Pathological Parsing**

To construct the anatomical and pathological representation core of EyeWorld, we built a large-scale biomedical image parsing dataset in which each image was paired with structured pixel-level segmentation masks and canonical semantic labels. Manual and semi-automatic annotations were verified by retinal specialists to ensure label accuracy and inter-observer consistency. The segmentation tasks span multiple retinal imaging modalities and encompass a broad range of anatomical structures and disease manifestations, including optic discs, retinal and subretinal layers, vascular networks, fluid accumulations, hyperfluorescent lesions and Choroidal neovascularization (CNV).

To enhance semantic grounding, we incorporated diverse natural-language prompts describing explicit visual goals, such as "highlight the optic disc using blue", "segment microaneurysms using cyan" and "outline the fluid". By aligning linguistic cues with visual representations and by jointly training segmentation and detection tasks within a unified framework, EyeWorld learns a coherent anatomical-pathological parsing of the retina. Each anatomical task was further paired with complementary mask-to-image and text-to-image training samples to encourage multi-task interdependence and joint reasoning across segmentation and detection.

To improve model generalization and robustness, we applied a targeted set of spatial and intensity augmentations. Spatial transformations included random image scaling. Intensity augmentations simulated real-world acquisition variability through brightness offsets of -10 and +20, contrast scaling within [0.8, 1.2] and gamma adjustments within the same range. In addition, contrast-limited adaptive histogram equalization (CLAHE) was used with a clip limit of 2.0 and a tile grid size of 8 × 8 to improve local contrast in fundus, angiographic and OCT images.

**Image Quality Enhancement**

Retinal fundus images often exhibit substantial quality variability due to inconsistent illumination, heterogeneous imaging devices and operator differences. The quality enhancement task was constructed by randomly sampling CFP images from the CFP collections described above, including both healthy and diseased cases, rather than introducing additional sources. To model real-world degradations, we implemented a degradation and restoration pipeline inspired by prior ophthalmic image restoration studies[28]. The pipeline introduced illumination imbalance, Gaussian blur, motion blur, random spot occlusions and their combinations. Downsampled versions of each image were generated with scale factors of 5, 7, 9, 11 and 13 to form paired super-resolution data. For completion tasks, inpainting and outpainting samples were automatically created using deterministic geometry-based masking implemented with the Pillow library. Inpainting applied one to five rectangular masks covering 10% to 50% of the image area. Outpainting retained an elliptical region covering 30% to 80% of the image and masked the surrounding context, prompting the model to extrapolate retinal structures beyond the visible field.

This process produced diverse and reproducible samples that improve EyeWorld's ability to reconstruct and restore degraded ophthalmic images. In total, we curated a restoration dataset of 3,120 degraded images covering illumination imbalance, blur, spot artefacts and their combinations. We further constructed a super-resolution dataset of 6,000 paired samples with downscaling factors of 5, 7, 9, 11 and 13. For completion tasks, we included inpainting and outpainting datasets containing 10,978 and 11,246 images, respectively. These data support restoration, super-resolution and missing-region reconstruction, which are used both as standalone generative tasks and as perturbation regimes to improve robustness.

**Progression Prediction**

To model disease progression and post-treatment structural recovery, EyeWorld incorporates progression-prediction tasks based on the longitudinal datasets described above. For the post-vitrectomy macular hole cohorts, inter-scan structural misalignment between preoperative and postoperative OCT volumes poses a particular challenge. To address this, we introduced an RPE-based alignment strategy. Specifically, a pretrained nnU-Net model was used to segment the retinal pigment epithelium (RPE) layer from postoperative OCT scans in the training set. The resulting binary masks were included as the second input image to provide position reference for retinal depth alignment. Importantly, the RPE mask serves as an alignment prior rather than an oracle for recovery. To further guide the model's focus, the textual instruction accompanying each sample was augmented with an explicit prompt: "Align the retinal position according to the RPE layer mask from image 2." This additional conditioning encourages EyeWorld to concentrate on the structural recovery of the macular hole region rather than learning absolute layer positions in postoperative OCTs. By incorporating anatomical priors into both the visual and textual inputs, the model learns to infer morphological healing patterns while remaining invariant to inter-patient positional variation.

**Exemplar-based conditioning protocol**

To improve cross-task generalization and reasoning consistency, we constructed a demonstration-conditioned instruction dataset inspired by the principles of in-context learning. For each existing task type, including segmentation, detection, translation and quality enhancement, we generated example-conditioned pairs by combining two related but distinct samples. For each such pair, one image-mask pair from the training corpus was used to form a demonstration image (the first input image) that illustrates the desired operation, such as delineating lesion boundaries, performing modality translation or restoring degraded regions. Another image (the second input image) served as the target that requires the same operation to be applied. The accompanying textual instruction was revised to explicitly describe this relationship, for example: "According to the demonstration shown in image 1, apply the same process to image 2." During data construction, the demonstration and target images were always selected from different patients to prevent identity or spatial leakage. This design encourages the model to learn the underlying transformation principle rather than memorizing patient-specific features. As illustrated conceptually in Fig. 8d, each demonstration-conditioned sample thus consists of a reference pair and a target image. The reference pair provides contextual information, such as anatomical structures or lesion patterns, while the target image requires the same operation to be applied. At inference time, users can provide a query image together with an annotated reference example, and the model infers the appropriate transformation by following the demonstrated operation.

## Model design and training details

Model architecture. EyeWorld adopts a unified multimodal generative architecture that separates semantic conditioning from image synthesis. Formally, let $x_{in}^t$ denote the input ophthalmic image at time $t$, and $\tau$ denote the textual instruction. For longitudinal forecasting, the temporal interval $\Delta t$ is provided implicitly as a natural-language phrase in $\tau$ (e.g., "Predict the post-operative retinal image at 6 months..."), enabling time-aware simulation without introducing extra numeric control variables. The model constructs a latent ocular state via two complementary pathways: a high-level semantic pathway for task specification and a low-level structural pathway for preserving patient-specific anatomy. First, a multimodal transformer $\Phi_M$ processes an interleaved sequence of text tokens from $\tau$ and ViT-encoded visual tokens from $x_t$, producing high-level multimodal hidden states $c_{sem}$ as semantic conditions:

$$c_{sem} = \Phi_M\left(T_{text}(\tau), T_{vit}(x_{in}^t)\right) \tag{1}$$

In parallel, to preserve fine-grained spatial structures required for medical imaging (e.g., vessel topology or lesion boundaries), a variational image tokenizer (VAE encoder) $E_{en}$ encodes the input image $x_{in}^t$ into a latent structural representation $z_{struc}$:

$$z_{struc} = E_{en}(x_{in}^t) \tag{2}$$

We define the latent ocular state as $S_t = (c_{sem}, z_{struc})$, unifying task semantics and patient-specific structure in a shared multimodal state space that supports multiple ophthalmic tasks through prompt variation. At the core of EyeWorld, an ophthalmic world simulator is implemented as a diffusion-based image decoder $F_{diff}$. For longitudinal forecasting, the model does not merely "edit" an image; it simulates an plausible follow-up transition $S_t \rightarrow S_{t+\Delta t}$, where temporal cues in the instruction $\tau$ guide the diffusion process to predict disease evolution consistent with clinical progression patterns. The final output $x_{out}$ is synthesized by the VAE decoder $E_{de}$, which projects the denoised target latent back into the pixel space. The target latent is generated by $F_{diff}$ by denoising random noise $\epsilon$ under joint conditioning on semantic states $c_{sem}$ and the structural latents $z_{struc}$:

$$x_{out} = E_{de}\left(F_{diff}\left(\epsilon, \underbrace{c_{sem}, z_{struc}}_{S_t}\right)\right) \tag{3}$$

As illustrated in Fig. 1c, from the latent ocular state $S_t$ (or its time-conditioned state $S_{t+\Delta t}$), EyeWorld performs three task-conditioned projections by varying the instruction content $\tau$, including (i) fine-grained anatomy and pathology projection, (ii) appearance and synthesis projection, and (iii) longitudinal forecasting projection. We initialize EyeWorld from a publicly released multimodal generative checkpoint and adapt it to ophthalmic tasks via supervised fine-tuning on paired medical images with pixel-level masks, modality-translation pairs, and longitudinal scan pairs; the architectural backbone is kept unchanged, while domain specialization is achieved through the curriculum-and-mining strategy described below.

Training strategy. Building on this initialization, to better accommodate long-tailed pathological patterns and heterogeneous acquisition conditions, we employed a multi-stage training strategy inspired by curriculum learning with explicit hard-example mining. We first fine-tuned on the full training corpus to establish coarse visual-language alignment and stable task behaviour. We then re-evaluated model predictions to identify systematically difficult samples. For segmentation and detection tasks, samples with Dice or mIoU below 0.5 were marked as hard examples, whereas for generative tasks, samples with SSIM below 0.5 were flagged. Hard examples were selectively re-augmented using lesion-level perturbations and global intensity transformations to improve robustness on rare lesions and failure modes. This design isolates where the foundation model fails in ophthalmic distributions and allocates additional supervision precisely to those failure modes, rather than uniformly increasing fine-tuning epochs. To balance computational cost and micro-lesion fidelity, the main training was conducted at 256×256 resolution, followed by a high-resolution adaptation stage at 512×512 to recover high-frequency details and improve micro-lesion localization.

All experiments were implemented in PyTorch (Python 3.11) and conducted on an NVIDIA A100-SXM4-80GB GPU cluster. The network was optimized using the AdamW optimizer ($\beta_1$=0.9, $\beta_2$=0.95, weight decay=0.01, $\varepsilon$=1×10$^{-8}$). The initial learning rate was set to 1×10$^{-5}$ and scheduled using a constant-with-warmup strategy, where the learning rate linearly increased from 1×10$^{-18}$ to its target value over the first 500 optimization steps and then remained constant throughout training. This approach stabilized early-stage convergence while maintaining consistent gradient updates in later stages. Gradient accumulation was employed to achieve an effective batch size of 64 across two NVIDIA A100-SXM4-80GB GPUs, and mixed-precision training was used to improve efficiency and numerical stability.

**Details of the comparison models**

We selected four recent general-purpose generative models as comparison baselines: Step1X-Edit, ChronoEdit, OmniGen2, and Qwen-Image, because they represent distinct design choices for instruction-following image transformation: (i) MLLM understanding + diffusion decoding with explicit token selection (Step1X-Edit), (ii) repurposing video priors to enforce physical coherence (ChronoEdit), (iii) a unified text–image generator that decouples autoregressive language modeling from diffusion-based image synthesis (OmniGen2), and (iv) a diffusion foundation model with a double-stream architecture and dedicated multimodal positional encoding for high-fidelity editing (Qwen-Image). Extended Data Table 1 summarize each model in terms of its core mechanism, conditioning pathway, and the capability it is intended to provide in our evaluation.

Step1X-Edit frames image editing as "MLLM understanding + diffusion decoding." Given an editing instruction and an image, a multimodal LLM (Qwen-VL) processes both jointly to capture instruction–image semantics, then discards system-prefix token embeddings to keep only embeddings that align with the edit information. These embeddings are passed through a lightweight connector to form a compact multimodal condition for a Diffusion Transformer (DiT) image decoder. The method also computes an average-pooled embedding that is projected into a global guidance vector to strengthen semantics for downstream diffusion. It is trained on more than one million high-quality instruction–image triplets covering 11 editing categories, enabling alignment with user-defined editing instructions.

ChronoEdit is an image-editing foundation model explicitly designed to preserve physical consistency by turning a pretrained video generative model into an instruction-guided image editor. Its key idea is to treat an edit as a temporally coherent transition from an input frame to an output frame, and to encourage coherent changes by inserting intermediate latent frames as "temporal reasoning tokens" between the input and the target. During training, standard image-editing pairs are reinterpreted as two-frame videos, while real or synthetic videos provide intermediate frames that act as reasoning tokens; this unifies learning from image pairs and full sequences within one formulation.

OmniGen2 is a multimodal generative model that supports both text generation and image generation in a single framework by decoupling autoregressive (text) modeling and diffusion-based (image) modeling. It leverages a foundational MLLM transformer initialized from Qwen2.5-VL-3B, and uses a special token ("<|img|>") to trigger an image-generation branch. When the image token is produced, the MLLM hidden states serve as conditional inputs to a dedicated diffusion decoder. The model also introduces a position-encoding design (Omni-RoPE) that combines a sequence or modality identifier with local 2D spatial coordinates to distinguish multiple images while encouraging spatial consistency during editing.

Qwen-Image is a diffusion-based image foundation model built around a standard double-stream Multimodal Diffusion Transformer (MMDiT). It uses a frozen Qwen2.5-VL as the multimodal condition encoder and a VAE as the image tokenizer, with a dedicated positional encoding scheme (MSRoPE) to jointly represent image and text positions. For image editing, Qwen-Image emphasizes both semantic coherence and visual consistency by encoding the input image into two complementary representations: semantic features extracted via Qwen-VL (high-level scene understanding) and reconstructive features from the VAE encoder (low-level visual details). Both are fed into the MMDiT as conditioning signals, enabling targeted modifications while preserving non-edited regions.

Together, these models span the spectrum from editing systems (Step1X-Edit, ChronoEdit) to unified generalist generators (OmniGen2, Qwen-Image), providing comprehensive baselines for assessing EyeWorld's capabilities across anatomical parsing, cross-modality translation, enhancement and progression prediction. We fine-tuned these models using identical hyperparameter setups to ensure a fair comparison. All baselines were fine-tuned on the same ophthalmic data splits as EyeWorld for each task family, with identical prompts, mask color conventions and post-processing. This ensures that performance differences reflect model design and learned representations rather than data.

**Evaluation metrics**

To comprehensively evaluate the performance of EyeWorld across tasks, we adopted quantitative metrics aligned with established standards in medical image analysis. The evaluation covers three major categories, including segmentation, detection, and generative tasks, ensuring fair and interpretable comparisons across imaging modalities.

For segmentation tasks, performance was evaluated using the Dice, mean intersection over union (mIoU), accuracy, precision, sensitivity, and specificity. The Dice score quantifies the spatial overlap between the predicted segmentation and the ground truth. The Dice score ranges from 0 to 1, where 1 indicates perfect overlap and 0 indicates no overlap. Accordingly, a higher Dice score reflects better segmentation performance and greater agreement with the reference annotation. The mIoU complements Dice by penalizing both over- and under-segmentation across classes. Accuracy, precision, sensitivity, and specificity jointly assess the model's ability to discriminate between pathological and non-pathological regions, providing a balanced evaluation of anatomical and lesion segmentation performance.

For detection and localization tasks, including lesion or object-level identification, evaluation was performed using mean intersection over union (mIoU), precision, recall, F1-score, and mean average precision (mAP). The mIoU quantifies the spatial agreement between predicted and reference bounding boxes, whereas precision and recall capture the balance between accuracy and completeness of detections. The F1-score reflects their harmonic mean, and mAP summarizes overall detection precision across multiple confidence thresholds.

For generation and image-quality tasks, including quality enhancement, modality translation, and post-treatment prediction, evaluation relied on perceptual and fidelity-based metrics: peak signal-to-noise ratio (PSNR), structural similarity index (SSIM), learned perceptual image patch similarity (LPIPS), Fréchet inception distance (FID), and inception score (IS)[27]. PSNR and SSIM quantify pixel-level and structural fidelity between generated and reference images, while LPIPS measures perceptual similarity in a deep feature space. FID evaluates both fidelity and diversity by computing the distributional distance between generated and real images in a pretrained Inception feature space. The Inception Score assesses image realism based on the entropy of class predictions. Collectively, these metrics provide a comprehensive assessment of generative quality from both perceptual and quantitative perspectives.

In addition to global image-level fidelity metrics, we visualize the direction and spatial distribution of predicted changes using a tissue-masked signed difference map. Unlike standard difference maps that can be confounded by non-retinal artifacts, our analysis is restricted to the biologically relevant region. Specifically, we first derive a tissue mask $M$ from the reference image and restrict all analyses to this retinal tissue region. Given a "before" image $x_{in}^{t}$ and an "after" image $x_{out}^{t+\Delta t}$, we compute signed change field

$$D_{sgn}(p)=mean_c\left(x_{in}^{t}(p)-x_{out}^{t+\Delta t}(p)\right)\cdot M(p) \qquad (4)$$

and normalize it by its maximum absolute value to obtain $\widetilde{D}_{sgn}\in[-1,1]$. We render $\widetilde{D}_{sgn}$ with a red-blue colormap, where red indicates intensity increase and blue indicates decrease, providing a spatially resolved view of the polarity and distribution of predicted changes (e.g., progression versus regression patterns). To facilitate consistent visual inspection, we additionally annotate the signed change map with an arrow pointing to the location of the maximum change magnitude within the tissue region. This automated focus annotation highlights the most prominent change location without affecting any quantitative metric. Together, the signed change map and focus annotation complement global metrics by characterizing the directionality and spatial concentration of structural deviations, which is particularly relevant for evaluating longitudinal progression and post-treatment recovery patterns.

**Data availability**

We do not have permission to redistribute the datasets used for developing EyeWorld, the data may be available under constrained access from the corresponding author upon reasonable request. Downstream datasets can be accessed via the links: PALM (https://doi.org/10.6084/m9.figshare.c.6224616.v1), IDRiD (https://ieee-dataport.org/open-access/indian-diabetic-retinopathy-image-dataset-idrid), STARE (https://cecas.clemson.edu/~ahoover/stare/), HRF (https://www5.informatik.uni-erlangen.de/research/data/fundus-images/), APTOS-2024 (https://doi.org/10.1016/j.media.2026.103942), MESSIDOR (https://www.adcis.net/en/third-party/messidor/), MMAC (https://zenodo.org/records/11025749), ODIR (https://www.kaggle.com/datasets/andrewmvd/ocular-disease-recognition-odir5k), OIMHS (https://www.nature.com/articles/s41597-023-02675-1#Tab1), MH cohort (preoperative and postoperative) (https://tvst.arvojournals.org/article.aspx?articleid=2778731), MARIO (https://arxiv.org/html/2506.02976v1#S4)

## Acknowledgement

We thank the InnoHK HKSAR Government for providing valuable support. The research work described in this paper was conducted in the JC STEM Lab of Innovative Light Therapy for Eye Diseases, funded by The Hong Kong Jockey Club Charities Trust. The study was supported by the Global STEM Professorship Scheme (P0046113) from HKSAR and Henry G. Leong Endowed Professorship in Elderly Vision Health.

## Author contributions

Conceptualization: DS, MH, ZG
Methodology: DS, ZG, XC
Formal analysis: ZG, XW, ZL
Resources: DS, MH
Writing - Original Draft: ZG, DS, ZL
Writing - Review & Editing: DS, XC, XW, ZL, RC, BL, BY, ZW, KJ, JY, YT
Visualization: ZG

Supervision: DS, MH
Funding acquisition: DS, MH

**Competing interests**

Prof. He is a medical officer of Optain Health Ltd. All other authors declare no competing interests.

**Additional information**

Supplementary Information is available for this paper. Correspondence and requests for materials should be addressed to danli.shi@polyu.edu.hk or mingguang.he@polyu.edu.hk.

**Extended Data Table**

Extended Data Table 1 | Comparison baselines and their conditioning pathways.

| Model | Step1X-Edit | ChronoEdit | OmniGen2 | Qwen-Image | EyeWorld |
|---|---|---|---|---|---|
| Backbone | MLLM + diffusion transformer (DiT) | Video diffusion model | Decoupled AR + Diffusion; VAE tokenizer | Dual-stream MMDiT; VAE tokenizer | Decoupled AR + Diffusion; VAE tokenizer |
| Primary emphasis | Instruction alignment | Physical/Temporal Coherence | Unified text↔image generation and editing | Region Preservation | Micro-structure fidelity |
| Semantic conditioning (high-level) | Filtered MLLM embeddings + Global guidance | "Temporal reasoning" at inference | MLLM hidden states (Omni-RoPE) | Multimodal alignment via MSRoPE | MLLM hidden states (Omni-RoPE) |
| Structure-preserving path (low-level) | VAE-derived visual tokens | Two-frame video formulation | VAE latents + position encoding | VAE reconstructive features + MLLM semantic features | VAE latents + position encoding + Curriculum Learning + Hard-Example Mining |